\documentclass[10pt]{article}
\usepackage[preprint]{tmlr}

\usepackage{hyperref}
\usepackage{orcidlink}
\usepackage{url}
\usepackage{graphicx}
\usepackage{makecell}
\usepackage{booktabs}
\usepackage{amsmath} 
\usepackage{multirow}
\usepackage{tikz}
\usepackage{subcaption}
\usepackage{natbib}
\usetikzlibrary{decorations.pathreplacing,patterns}
\usetikzlibrary{positioning, arrows.meta, shapes.geometric, fit, backgrounds, calc}

\definecolor{lightgreen}{RGB}{200, 230, 200}
\definecolor{lightblue}{RGB}{173, 216, 230}
\definecolor{lighterblue}{RGB}{211, 225, 245}
\definecolor{lightpink}{RGB}{245, 200, 195}

\title{EEG-CLIP: Learning EEG representations from natural language descriptions}

\author{\name Tidiane Camaret Ndir \orcidlink{0009-0009-9523-2157} \email tidiane.camaret.ndir@uniklinik-freiburg.de \\
      \addr Medical Physics, Department of Diagnostic and Interventional Radiology\\
      Medical Center—University of Freiburg, Faculty of Medicine\\
      University of Freiburg, Germany
      \AND
      \name Robin T. Schirrmeister \orcidlink{0000-0002-5518-7445} \email robin.schirrmeister@uniklinik-freiburg.de \\
      \addr Neuromedical A.I. Lab, Department of Neurosurgery\\
      Medical Center—University of Freiburg, Faculty of Medicine\\
      University of Freiburg, Germany
      \AND
      \name Tonio Ball \orcidlink{0000-0002-5518-7445} \email tonio.ball@uniklinik-freiburg.de \\
      \addr BrainLinks-BrainTools, IMBIT\\
      University of Freiburg, Germany
      }

\begin{document}
\maketitle

\begin{abstract}
Deep networks for electroencephalogram (EEG) decoding are often only trained to solve one specific task, such as pathology or age decoding. A more general task-agnostic approach is to train deep networks to match a (clinical) EEG recording to its corresponding textual medical report and vice versa. This approach was pioneered in the computer vision domain matching images and their text captions and subsequently allowed to do successful zero-shot decoding using textual class prompts. In this work, we follow this approach and develop a contrastive learning framework, EEG-CLIP, that aligns the EEG time series and the descriptions of the corresponding clinical text in a shared embedding space. We investigated its potential for versatile EEG decoding, evaluating performance in a range of few-shot and zero-shot settings. Overall, we show that EEG-CLIP manages to non-trivially align text and EEG representations. Our work presents a promising approach to learn general EEG representations, which could enable easier analyses of diverse decoding questions through zero-shot decoding or training task-specific models from fewer training examples. The code for reproducing our results is available at \url{https://github.com/tidiane-camaret/EEGClip}.
\end{abstract}

{\bf Keywords:} Electroencephalogram (EEG), Contrastive Learning, Multimodal Representation, Zero-shot Classification, Clinical Text Processing, Neural Time Series, Transfer Learning

\section{Introduction}

Recent advances in machine learning have led to deep neural networks being commonly applied to electroencephalogram (EEG) data for a variety of decoding tasks \citep{roy2019}. While deep learning models can achieve state-of-the-art performance on specialized EEG tasks, their learned representations can often only be used for one specific task. Most EEG analyses focus on training task-specific models for one type of classification or regression problem \citep{heilmeyer2018}. However, medical EEG recordings are often accompanied by rich unstructured annotations in the form of free text reports written by neurologists and medical experts—a potentially valuable source of supervision that remains largely untapped.

In the computer vision domain, Contrastive Language–Image Pre-training or CLIP \citep{radford2021} leverages text-image pairing to learn visual representations that effectively transfer across tasks. CLIP has demonstrated remarkable zero-shot generalization capabilities by learning to align images with natural language descriptions, enabling classification of previously unseen categories and adaptation to novel visual tasks without additional training.

Inspired by CLIP, we propose EEG-CLIP : a contrastive learning approach to align EEG time-series data with corresponding clinical text descriptions in a shared embedding space. This work explores two central questions: (i) how clinical text reports can be effectively incorporated into EEG representation learning, and (ii) whether this multimodal approach enables more generalizable representations that transfer across diverse EEG decoding tasks.

We demonstrate EEG-CLIP's potential for versatile EEG decoding through extensive evaluation on few-shot and zero-shot learning tasks. Our results show that EEG-CLIP achieves strong zero-shot classification performance and consistently outperforms previous transfer learning approaches and task-specific models when labeled data are scarce. This presents a promising direction for EEG analysis by enabling zero-shot inference through natural language queries and more efficient training of specialized models with limited annotations.

\textbf{Remark} Recently and after the completion of the study presented in this manuscript, \citep{Gijsen2025} also proposed EEG-language models that align EEG data with clinical reports for pathology detection. Their work explores multiple alignment strategies, including a multiple instance learning extension for flexible matching between EEG segments and text portions. Their approach is primarily focused on pathology detection and classification of epileptiform activity, while our EEG-CLIP study examines model performance on diverse decoding objectives including age, gender, and medication prediction, providing further insights into the versatility of language-supervised EEG representations.

\section{Related Work}
\subsection{Deep-Learning based EEG decoding}

Deep learning has revolutionized EEG analysis by enabling end-to-end decoding directly from raw signals without hand-crafted features. Convolutional neural networks (CNNs) have shown particular promise, with recent advances like batch normalization and exponential linear units boosting performance to match or exceed traditional methods like filter bank common spatial patterns (FBCSP) \citep{schirrmeister2018} These architectures automatically learn hierarchical representations that capture relevant spectral and spatial patterns in EEG data.

Various neural network architectures have been applied to EEG tasks, from shallow CNNs for efficient processing to recurrent networks for capturing temporal dependencies. Recent comparative studies have demonstrated that specialized deep learning models can outperform traditional approaches  on standard benchmarks such as BCI Competition datasets. Beyond classification, newer approaches like EEG-to-text decoding leverage advanced neural architectures combined with probabilistic modeling to translate neural activity into human-readable text \citep{levy2025}, expanding the potential applications for EEG-based interfaces.

Multimodal approaches have also shown promise, such as \cite{khan_early_2018} who combined EEG with fNIRS to capture complementary neurophysiological signals. While they used traditional signal processing rather than deep learning, their work highlights the value of integrating EEG with additional information sources, a principle that motivates our EEG-text alignment approach.

While task-specific models dominate current approaches, our work explores a more general representation learning framework that leverages the rich information in clinical text reports to develop versatile EEG embeddings useful across multiple decoding tasks.

\subsection{Contrastive Learning for Multimodal Alignment}

Self-supervised contrastive learning has recently emerged as a powerful approach to learning general visual representations. Models like CLIP are trained to align the image embeddings $x_i$ and the corresponding text embeddings $y_i$ by minimizing contrast loss $\mathcal{L}$:

$$\mathcal{L}=\sum_{i=1}^{N}-\log \frac{\exp(\text{sim}(x_i,y_i)/\tau)}{\sum_{j=1}^{N}\exp(\text{sim}(x_i,y_j)/\tau)}$$

where $sim(.)$ is a measure of similarity. This objective brings the matching image-text pairs closer and separates the mismatched pairs in the learned embedding space.

CLIP was trained on a large dataset of 400 million image-text pairs from diverse Internet sources with unstructured annotations. Through this natural language supervision, CLIP developed versatile image representations that achieve strong zero-shot inference on downstream tasks by querying the aligned embedding space.

The success of CLIP highlights the promise of contrastive learning approaches and the use of readily available text data to learn transferable representations of other modalities.

\section{Methods}

\subsection{Dataset}

The Temple University Hospital EEG corpus \citep{obeid2016} contains over 25,000 EEG recordings collected from over 14,000 patients between 2002 and 2015. The large number of EEG recordings make this a valuable training dataset for deep learning models to learn to decode  information such as pathology or age from the EEG and be able to generalize to unseen EEG recordings. The TUH Abnormal dataset (TUAB) is a demographically balanced subset with binary labels indicating pathological or nonpathological diagnosis of each recording. It is partitioned into training (1,387 normal and 1,398 abnormal files) and evaluation (150 normal and 130 abnormal files) sets. It contains a variety of pathological conditions.

Each recording contains additional labels : “age” (integer), “gender” (“M” or “F”), and “report” (string), a medical report written in natural language. The report is divided in 15 sections, listed in Table~\ref{tuab-report-stats}.

\begin{table}[t]
\caption{\textbf{Report section characteristics in the TUAB dataset.} Distribution of medical report sections showing highest coverage in diagnostic sections (Impression, Description, Clinical History) with substantial text content, versus limited coverage in specialized sections.}
\label{tuab-report-stats}
\centering
\renewcommand{\arraystretch}{1.2}
\begin{tabular}{l r r}
\toprule
\textbf{Record section} & \textbf{Non-empty entries} & \textbf{Average word count} \\
\midrule
IMPRESSION & 2,971 & 16 \\
DESCRIPTION OF THE RECORD & 2,964 & 70 \\
CLINICAL HISTORY & 2,947 & 26 \\
MEDICATIONS & 2,893 & 4 \\
INTRODUCTION & 2,840 & 31 \\
CLINICAL CORRELATION & 2,698 & 31 \\
\midrule
HEART RATE & 1,458 & 2 \\
FINDINGS & 887 & 16 \\
REASON FOR STUDY & 713 & 2 \\
TECHNICAL DIFFICULTIES & 684 & 3 \\
EVENTS & 569 & 8 \\
\midrule
CONDITION OF THE RECORDING & 116 & 30 \\
PAST MEDICAL HISTORY & 19 & 8 \\
TYPE OF STUDY & 16 & 3 \\
ACTIVATION PROCEDURES & 9 & 3 \\
\bottomrule

\end{tabular}
\end{table}

\subsection{EEG data preprocessing}
We preprocess the EEG data, following the preprocessing steps from \cite{schirrmeister2018} :

\begin{itemize}
    \item Select a subset of 21 electrodes present in all recordings.
    \item Exclude the first minute of the recordings, and only use the first 2 minutes after that
    \item Clip the amplitude values to the range of ±800 \(\mu\) V to reduce the effects of strong artifacts.
    \item Resample the data to 100 Hz to further speed up the computation.
    \item Divide by 30 to get closer to unit variance
\end{itemize}

\subsection{Architecture and training details}

\begin{figure}[h]
\centering
\includegraphics[width=0.9\textwidth]{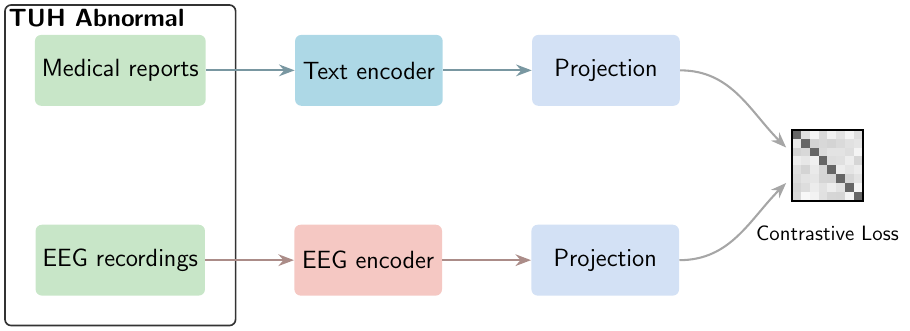} 
\caption{\textbf{Model architecture of EEG-CLIP.} The figure illustrates the dual-encoder architecture with an EEG encoder processing EEG time series data and a text encoder (pretrained BERT-based model) processing clinical reports. Both modalities are projected into a shared 64-dimensional embedding space through MLP projection heads. The contrastive loss optimizes for alignment between matching EEG-text pairs while pushing non-matching pairs apart, enabling the model to learn cross-modal representations that capture semantic relationships between neurophysiological patterns and their clinical descriptions.}
\label{fig:model_diagram}
\end{figure}

The EEG-CLIP model is composed of two main components: an EEG encoder and a text encoder. These encoders are designed to process EEG recordings and medical reports respectively, as depicted in Figure \ref{fig:model_diagram}.

For the EEG encoder, we use a convolutional neural network (CNN), Deep4 \citep{schirrmeister2018}, whose architecture is optimized for the classification of EEG data. The Deep4 Network features four convolution-max-pooling blocks, using batch normalization and dropout, followed by a dense softmax classification layer. This enables the model to learn hierarchical spatial-temporal representations of the EEG signal. The output is flattened and passed to a fully-connected layer to derive a 128-dimensional embedding.

For the text encoder, we leverage pretrained text encoders based on the BERT architecture \citep{devlin2019}. 
Such transformer-based models have shown state-of-the-art performance on a variety of natural language processing tasks. The advantage of these pretrained models is that they provide rich linguistic representations that can be effectively transferred to downstream tasks through finetuning. 

The EEG and text embeddings are then fed into MLP projection heads, consisting of 3 fully-connected layers with ReLU activations. The final layer outputs a 64-dimensional projection of the embedding for contrastive learning. This architecture allows the model to learn alignments between EEG windows and corresponding medical report sentences in a shared embedding space. The contrastive loss enables the useful semantic features to be captured.

We train EEG-CLIP using the Adam optimizer with a learning rate of $5 \times 10^{-3}$ and weight decay of $5 \times 10^{-4}$. The model is trained for 20 epochs with a batch size of 64. We use the same training/testing split as in the TUAB dataset. Each recording is split in windows of length 1200, corresponding to a 12 second period, and with a stride of 519, which ensures all timesteps are predicted without any gap by our Deep4 model.

\subsection{Evaluation methods}

Unlike models trained for a specific downstream task, EEG-CLIP has to learn broadly useful representations that capture semantic relationships between EEG signals and text. As such, evaluation methods must aim to quantify the general quality and transferability of the learned representations. 

Using the labels and medical reports provided in the TUAB dataset, we select 4 decoding tasks: 

\begin{itemize}
    \item “Pathological” : decode whether the recording was diagnosed as normal or pathological
    \item “Age”: decode whether the age of the patient is smaller or equal, or greater than 50
    \item “Gender”: decode the declared gender of the patient
    \item “Medication” : decode whether the medical report contains at least one of the 3 most common anticonvulsant medications ("keppra", "dilantin" and "depakote")
\end{itemize}

We then design multiple methods to evaluate the model on, listed in the following.

\subsubsection{Classification}

We compare EEG-CLIP against two baseline models to contextualize its performance:

\begin{itemize}
\item \textbf{Task-specific model (upper bound)}: A Deep4 CNN trained end-to-end from random initialization directly on each target task. This provides an upper bound since the entire model can optimize specifically for the task.

\item \textbf{Alternative task transfer model (lower bound)}: A Deep4 CNN first trained from random initialization on an unrelated task (e.g., age classification), then transferred to the target task (e.g., pathology detection) by freezing the encoder and training only a new classification head. This tests whether simple transfer learning from any EEG task provides useful features.
\end{itemize}

Both baselines use the same architecture as EEG-CLIP's EEG encoder but differ in their training approach: EEG-CLIP uses contrastive learning with text supervision, while baselines use standard supervised learning with task labels. Figure \ref{fig:baselines_schema} illustrates the different training strategies. 

\begin{figure}[h]
\centering
\scalebox{0.90}{
\begin{tikzpicture}[
    box/.style={rectangle, rounded corners=3pt, minimum height=0.8cm, minimum width=2cm, font=\sffamily\small, align=center},
    encoder/.style={box, fill=lightpink},
    classifier/.style={box, fill=lightgray},
    text_enc/.style={box, fill=lightblue},
    frozen/.style={box, fill=gray!30, pattern=north east lines, pattern color=gray!50},
    arrow/.style={
        ->, 
        thick,
        line width=1pt,
        draw=#1 
    },
    title/.style={font=\sffamily\bfseries},
    subtitle/.style={font=\sffamily\small, align=center}
]

    \node[subtitle] at (3.5, -0.8) {Step 1: Pre-training/Training};
    \node[subtitle] at (10.5, -0.8) {Step 2: Target Task Training};
    
    \node[title, align=center] at (-1.5, -2.5) {EEG-CLIP};
    
    \node[text_enc] (text1) at (2, -2) {Text\\Encoder};
    \node [encoder] (eeg1) at (2, -3.5) {EEG\\Encoder};
    \node (text_input1) at (0, -2) {Text};
    \node (eeg_input1) at (0, -3.5) {EEG};
    \draw[arrow] (eeg_input1) -- (eeg1);
    \draw[arrow] (text_input1) -- (text1);
    
    \node[align=center] (contrast) at (5, -2.75) {Contrastive\\Loss};
    \draw[arrow] (eeg1) -- (contrast);
    \draw[arrow] (text1) -- (contrast);
    
    \node[frozen] (eeg1_frozen) at (9, -2.5) {EEG\\Encoder\\(Frozen)};
    \node[classifier] (clf1) at (11.5, -2.5) {Classif.\\Head};
    \node (eeg_input1b) at (7, -2.5) {EEG};
    \draw[arrow] (eeg_input1b) -- (eeg1_frozen);
    \draw[arrow] (eeg1_frozen) -- (clf1);
    \node[align=center] (task1) at (14, -2.5) {Target\\Task};
    \draw[arrow] (clf1) -- (task1);
    
    \node[title, align=center] at (-1.5, -5.5) {Alternative\\Task\\Transfer};
    
    \node[encoder] (eeg2) at (2, -5.5) {EEG\\Encoder};
    \node[classifier] (clf2a) at (4.5, -5.5) {Classif.\\Head};
    \node (eeg_input2) at (0, -5.5) {EEG};
    \draw[arrow] (eeg_input2) -- (eeg2);
    \draw[arrow] (eeg2) -- (clf2a);
    \node[subtitle] (irr_task) at (4.5, -4.8) {(e.g., Age)};
    
    \node[frozen] (eeg2_frozen) at (9, -5.5) {EEG\\Encoder\\(Frozen)};
    \node[classifier] (clf2b) at (11.5, -5.5) {New\\Classif.\\Head};
    \node (eeg_input2b) at (7, -5.5) {EEG};
    \draw[arrow] (eeg_input2b) -- (eeg2_frozen);
    \draw[arrow] (eeg2_frozen) -- (clf2b);
    \node[align=center] (task2) at (14, -5.5) {Target\\Task};
    \draw[arrow] (clf2b) -- (task2);
    
    \node[title, align=center] at (-1.5, -8) {Task-\\Specific};
    
    \node[encoder] (eeg3) at (2, -8) {EEG\\Encoder};
    \node[classifier] (clf3) at (4.5, -8) {Classif.\\Head};
    \node (eeg_input3) at (0, -8) {EEG};
    \draw[arrow] (eeg_input3) -- (eeg3);
    \draw[arrow] (eeg3) -- (clf3);
    \node[align=center] (task3) at (6.5, -8) {Target\\Task};
    \draw[arrow] (clf3) -- (task3);
    
    \node[subtitle] at (10.5, -8) {(End-to-end training on target task only)};
    
    \draw[dashed, gray] (6.5, -1.2) -- (6.5, -9);
    
\end{tikzpicture}
}

\caption{\textbf{Training approaches comparison.} Three methods for EEG classification: (1) EEG-CLIP uses contrastive pre-training with text before target task fine-tuning with frozen encoder, (2) Alternative Task Transfer pre-trains on an unrelated task then adapts to target task with frozen encoder, (3) Task-Specific trains directly on target task end-to-end.}
\label{fig:baselines_schema}
\end{figure}

\subsubsection{Zero-shot classification}

We also perform zero-shot evaluation, using the embeddings of class-specific text prompts as class prototypes for the trained EEG-CLIP model. For a given classification task, we define a typical prompt sentence for each class (see Table~\ref{prompts-zero-shot}) and calculate the distance of an EEG recording to those sentences in the shared embedding space. This allows us to measure the classification performance of EEG-CLIP without any training on the classification task labels.

\begin{table}[h]
\caption{\textbf{Text prompts for zero-shot classification.}
Concise natural language prompts representing each class for four classification tasks, enabling classification through EEG-text similarity without task-specific training.}
\label{prompts-zero-shot}
\centering
\renewcommand{\arraystretch}{1.3}
\begin{tabular}{l p{5.5cm} p{5.5cm}}
\toprule
\textbf{Task} & \textbf{Prompt A} & \textbf{Prompt B} \\
\midrule
Pathological & ``This is a normal recording'' & ``This is an abnormal recording'' \\
Age & ``The patient is under 50 years old'' & ``The patient is over 50 years old'' \\
Gender & ``The patient is male'' & ``The patient is female'' \\
Medication & ``No anti-epileptic drugs were prescribed to the patient'' & ``Anti-epileptic drugs were prescribed to the patient'' \\
\bottomrule
\end{tabular}
\end{table}

\newpage
\subsubsection{Classification in a low-data regime}

To further evaluate the generalization capability of the learned representations, we assess few-shot performance by training classifiers on varying fractions of a small labeled dataset. Specifically:

\begin{itemize}
    \item We hold out 20\% of the TUAB training set exclusively for few-shot training (never seen during EEG-CLIP's contrastive training)
    \item From this 20\% held-out set, we create subsets of sizes: 1/2, 1/5, 1/10, 1/20, and 1/50
\end{itemize}

All models (EEG-CLIP, task-specific, and alternative task) are trained on these identical data subsets. The key difference is that EEG-CLIP uses representations learned from contrastive training on the separate 60\% split, while the task-specific model trains from scratch. This ensures a fair comparison where all models have access to the same limited labeled data, isolating the benefit of pre-training. Figure \ref{fig:data-partition} illustrates the data partitioning strategy.

\begin{figure}[h]
\centering
\begin{tikzpicture}[
    box/.style={rectangle, draw, minimum height=0.8cm, font=\sffamily},
    section/.style={font=\sffamily\bfseries}
]
    
    \node[section] at (6,-0.5) {TUAB Official Dataset Split};
    
    \draw[fill=gray!10, draw=black] (0,-1) rectangle (12,-2);
    \draw[fill=gray!20, draw=black] (0,-1) rectangle (9.6,-2);
    
    \node at (4.8,-1.5) {Train (80\%)};
    \node at (10.8,-1.5) {Eval (20\%)};
    
    \node[section] at (6,-3) {Standard Classification Setup};

    \draw[fill=gray!30, draw=black] (0,-5) rectangle (9.6,-5.8);
    \draw[fill=blue!20, draw=black] (9.6,-5) rectangle (12,-5.8);
    
    \node at (4.8,-5.4) {Contrastive Training + Task Training};
    \node at (10.8,-5.4) {Task Eval};
    
    \node[section] at (6,-6.5) {Few-shot Classification Setup};
    
    \draw[fill=gray!10, draw=black] (0,-7) rectangle (12,-8.2);
    \draw[fill=gray!20, draw=black] (0,-7) rectangle (7.2,-8.2);
    \draw[fill=yellow!20, draw=black] (7.2,-7) rectangle (9.6,-8.2);
    \draw[fill=blue!20, draw=black] (9.6,-7) rectangle (12,-8.2);
    
    \node at (3.6,-7.4) {Contrastive};
    \node at (3.6,-7.8) {Training (60\%)};
    
    \node at (8.4,-7.4) {Task};
    \node at (8.4,-7.8) {Train. (20\%)};
    
    \node at (10.8,-7.4) {Task};
    \node at (10.8,-7.8) {Eval (20\%)};
    
    \draw [decorate,decoration={brace,amplitude=5pt,mirror}] (7.2,-8.2) -- (9.6,-8.2) 
      node[midway,below=8pt,font=\sffamily\small] {Varying fractions: $\frac{1}{2}$, $\frac{1}{5}$, $\frac{1}{10}$, $\frac{1}{20}$, $\frac{1}{50}$};
    
\end{tikzpicture}
\caption{Experimental data partitioning strategies for EEG-CLIP. The top section shows the official TUAB dataset split. The middle section illustrates the standard classification setup where the training portion is used for contrastive learning between EEG signals and text descriptions. The bottom section visualizes the few-shot learning approach: 60\% is used for EEG-CLIP's contrastive pre-training (without task labels), 20\% serves as the few-shot training set (from which varying fractions are sampled), and 20\% is held for evaluation. \textbf{In few-shot experiments, all compared models use only the same subsets from the 20\% Task Train split.}}
\label{fig:data-partition}
\end{figure}

\section{Results}

\subsection{Evaluation of the learned representations}

In this section, we present results evaluating the learned representations from EEG-CLIP across a diverse set of experiments. As a reminder, our evaluation methodology consisted of classification tasks using the full TUAB dataset, zero-shot classification using text prompts, and few-shot classification on a held-out dataset.

\subsubsection{Classification performance}

Table \ref{classif-acc-table} shows EEG-CLIP's classification performance across four tasks. With logistic regression, EEG-CLIP achieves balanced accuracies of 0.826 for pathological status, 0.713 for age, and 0.687 for gender. A 3-layer MLP classifier further improves results to 0.847, 0.747, and 0.702 respectively, indicating non-linear relationships in the embedding space.
The performance gap between EEG-CLIP+MLP and task-specific models remains small (0.004 for pathological, 0.039 for age, 0.050 for gender) despite the latter's end-to-end optimization advantage. Most importantly, EEG-CLIP consistently outperforms alternative task pretraining by 10.6\% for pathological, 6.2\% for age, and 3.5\% for gender classification. These quantitative results demonstrate that text-EEG contrastive learning produces more transferable representations than single-task supervised learning.

\begin{table}[h]
\caption{\textbf{Classification performance comparison (balanced accuracy).} EEG-CLIP approaches task-specific performance while substantially outperforming alternative task pretraining, demonstrating effective text-supervised representation learning.}
\label{classif-acc-table}
\centering
\renewcommand{\arraystretch}{1.2}
\begin{tabular}{l cccc}
\toprule
\multirow{2}{*}{\textbf{Task}} & \multicolumn{2}{c}{\textbf{EEG-CLIP}} & \multirow{2}{*}{\textbf{Task-specific}} & \multirow{2}{*}{\textbf{Alternative task}} \\
\cmidrule(lr){2-3}
 & \textbf{LogReg} & \textbf{MLP} & & \\
\midrule
Pathological & 0.826 & 0.847 & \textbf{0.851} & 0.741 \textit{(age)} \\
Age & 0.713 & 0.747 & \textbf{0.786} & 0.685 \textit{(pathological)} \\
Gender & 0.687 & 0.702 & \textbf{0.752} & 0.667 \textit{(pathological)} \\
Medication & 0.633 & 0.615 & \textbf{0.685} & 0.573 \textit{(pathological)} \\
\bottomrule
\end{tabular}
\end{table}

\newpage
\subsubsection{Zero-shot classification performance}

For zero-shot classification, we evaluate EEG-CLIP's ability to classify EEG recordings without any task-specific training. We compute similarities between EEG embeddings and text prompts (see Table~\ref{prompts-zero-shot}) in the shared embedding space. As shown in Table~\ref{zero-shot-classif-acc-table}, EEG-CLIP achieves remarkable zero-shot performance on the pathological task (0.755), demonstrating strong alignment between EEG signals and their textual descriptions. Performance on age classification (0.642) is also substantially above chance, while gender (0.567) and medication (0.532) show more modest scores. These results are particularly encouraging as they represent classification without any labeled training data, relying solely on the semantic alignment learned during contrastive training. The strong pathology detection performance suggests that diagnostic language in the medical reports is effectively aligned with corresponding neurological patterns in the EEG signals. This zero-shot capability could be especially valuable in clinical settings where labeled data for new tasks is scarce or unavailable.

\begin{table}[h]
\caption{\textbf{Zero-shot classification performance (balanced accuracy.)}
EEG-CLIP achieves strong performance for pathology detection (0.755) and age classification (0.642) using only text prompts, demonstrating effective EEG-text alignment.}
\label{zero-shot-classif-acc-table}
\centering
\renewcommand{\arraystretch}{1.2}
\begin{tabular}{l c}
\toprule
\textbf{Task} & \textbf{Accuracy} \\
\midrule
Pathological & 0.755 \\
Age & 0.642 \\
Gender & 0.567 \\
Medication & 0.532 \\
\bottomrule
\end{tabular}
\end{table}

\subsubsection{Few-shot classification performance}

On the pathological task, EEG-CLIP achieves 0.710 balanced accuracy on the held-out set. This approaches the 0.781 performance of a model trained from scratch with the same limited data. For age classification, EEG-CLIP even outperforms the specialized model. The medication task proves most challenging in the few-shot setting. However, all models struggle to exceed 0.6 accuracy, suggesting intrinsic difficulty of the binary prediction from small samples. The detailed results are presented in Table \ref{classif-few-shot-acc-table}.

\begin{table}[h]
\caption{\textbf{Few-shot learning performance (balanced accuracy).}
EEG-CLIP outperforms models trained from scratch on age classification, demonstrating representation transferability when labeled data is scarce.}
\label{classif-few-shot-acc-table}
\centering
\renewcommand{\arraystretch}{1.2}
\begin{tabular}{l ccc}
\toprule
\textbf{Task} & \textbf{EEG-CLIP + MLP} & \textbf{Task-specific} & \textbf{Alternative task} \\
\midrule
Pathological & 0.710 & \textbf{0.781} & 0.531 \textit{(age)} \\
Age & \textbf{0.712} & 0.621 & 0.631 \textit{(pathological)} \\
Gender & 0.550 & \textbf{0.648} & 0.512 \textit{(pathological)} \\
Medication & 0.551 & 0.575 & \textbf{0.598} \textit{(pathological)} \\
\bottomrule
\end{tabular}
\end{table}

Critically, EEG-CLIP substantially outperforms models pretrained on alternative tasks across all but one experiment. This demonstrates the concrete value of pretraining on aligned data, even when fine-tuning data is scarce.

\begin{figure}[h]
    \centering
    \begin{minipage}[b]{0.45\textwidth}
        \centering
        \includegraphics[width=\linewidth]{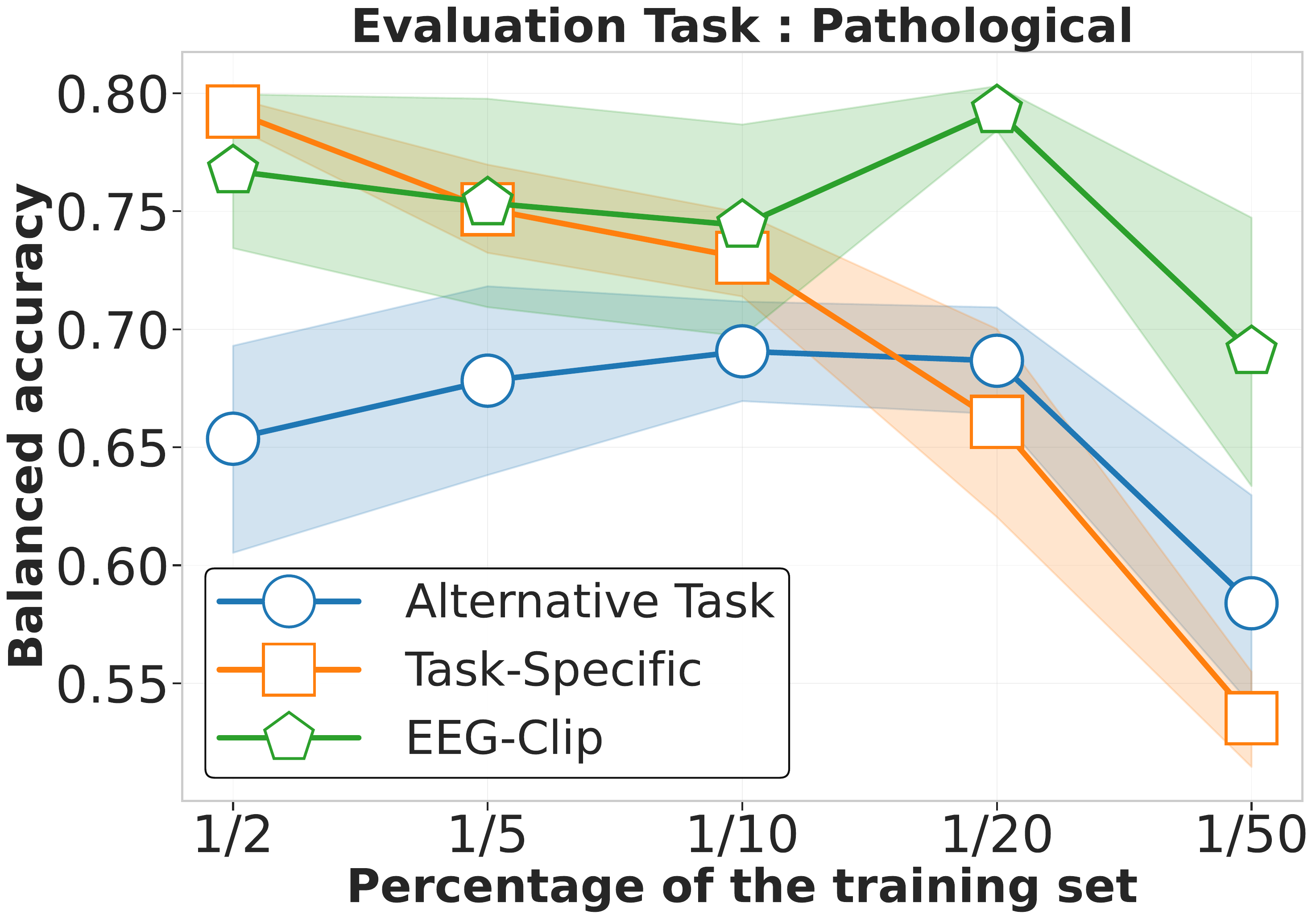}
        \label{fig:subfig1}
    \end{minipage}
    \hfill
    \begin{minipage}[b]{0.45\textwidth}
        \centering
        \includegraphics[width=\linewidth]{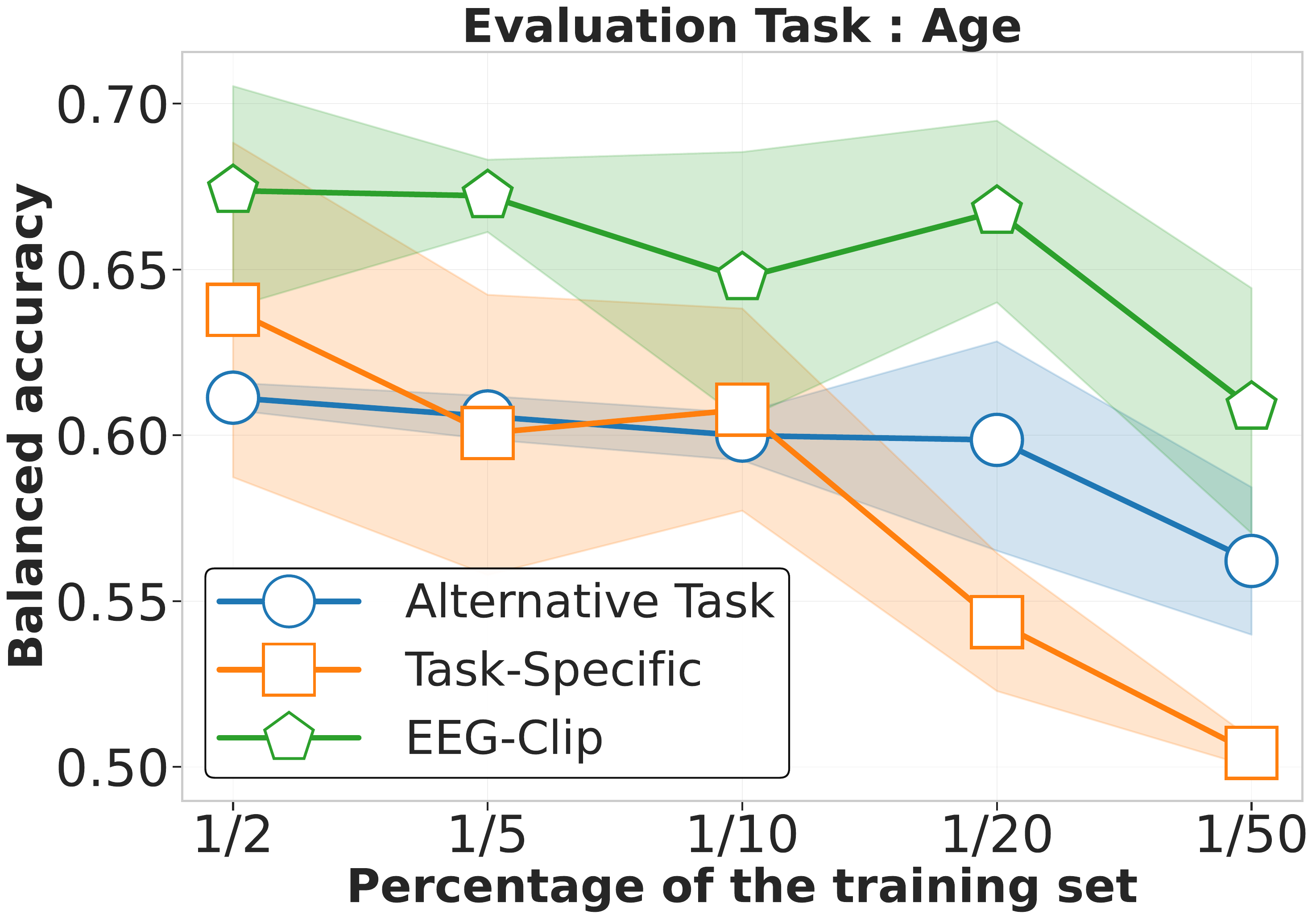}
        \label{fig:subfig2}
    \end{minipage}
    
    \vspace{0.5cm}
    
    \begin{minipage}[b]{0.45\textwidth}
        \centering
        \includegraphics[width=\linewidth]{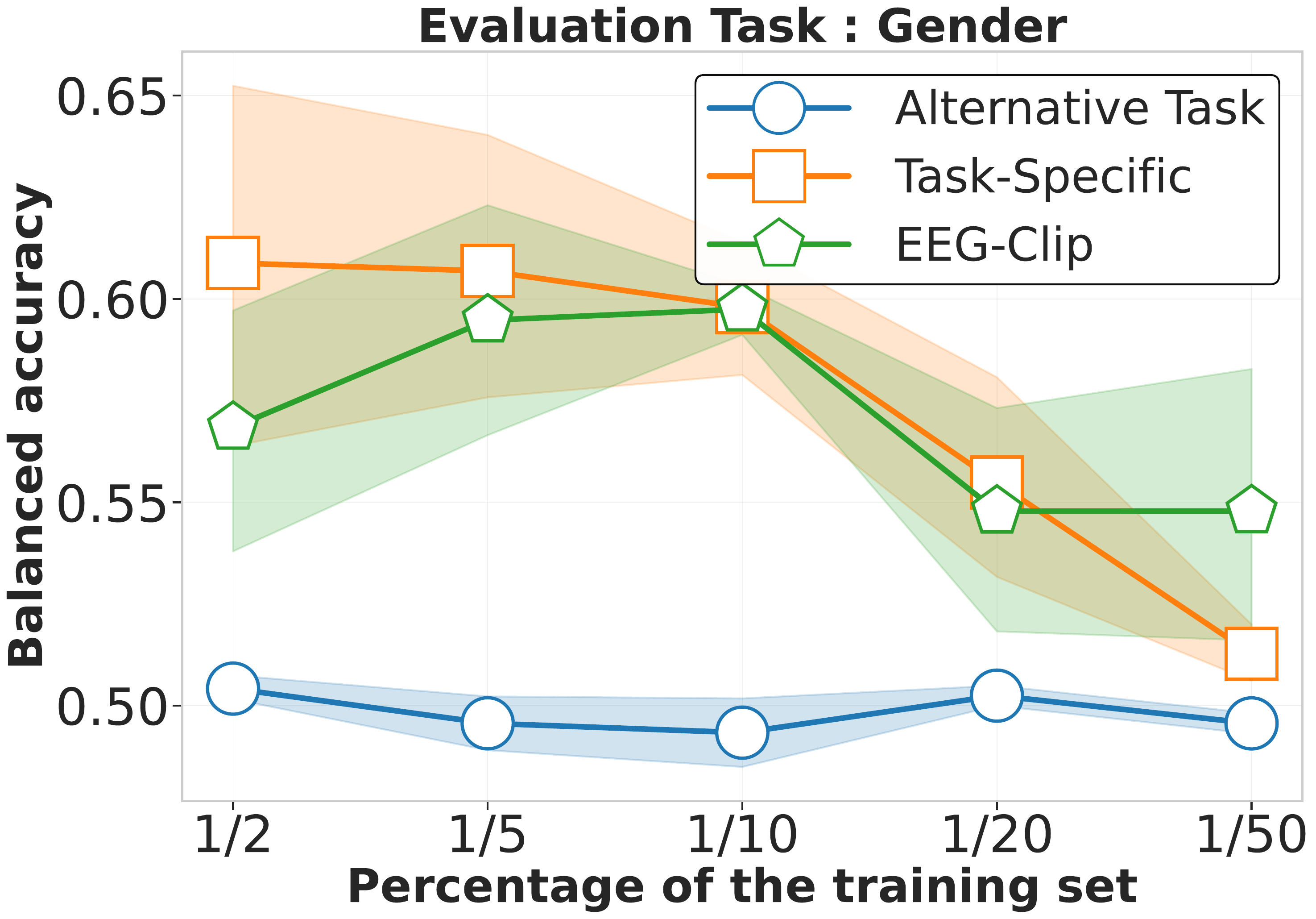}
        \label{fig:subfig3}
    \end{minipage}
    \hfill
    \begin{minipage}[b]{0.45\textwidth}
        \centering
        \includegraphics[width=\linewidth]{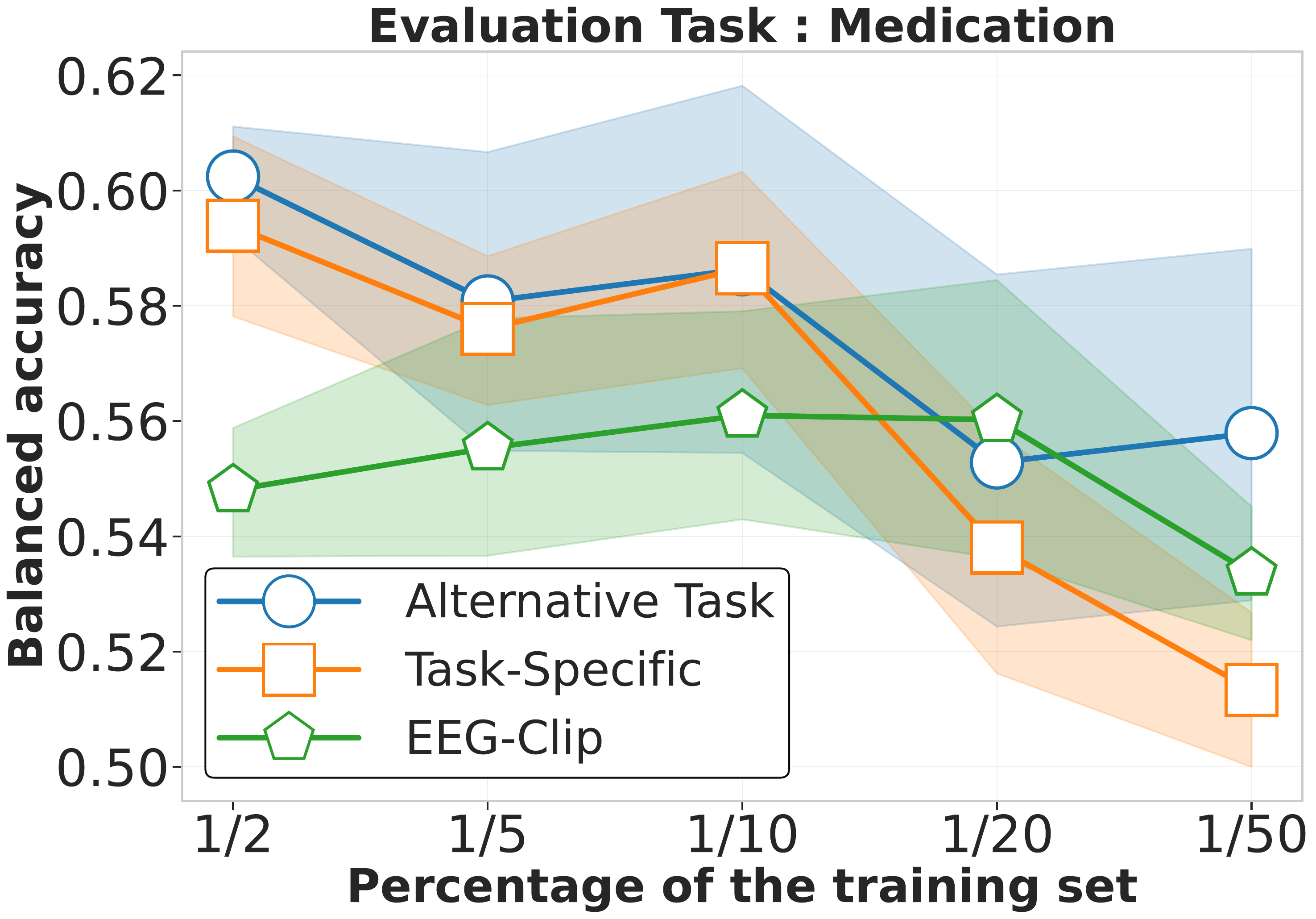}
        \label{fig:subfig4}
    \end{minipage}
    
    \caption{\textbf{Performance comparison in low-data regimes across four decoding tasks.} Each panel shows balanced accuracy as a function of training set size (from $\frac{1}{2}$ to $\frac{1}{50}$ of the full dataset) for three approaches: EEG-CLIP (green), task-specific models trained from scratch (orange), and models pretrained on alternative tasks (blue). For pathology detection (top left), EEG-CLIP maintains strong performance even with minimal data ($\frac{1}{20}$), outperforming both baselines as data becomes scarce. Age classification (top right) shows EEG-CLIP consistently outperforming other approaches across all data regimes. For gender (bottom left) and medication (bottom right) tasks, all models show performance degradation with reduced data, but EEG-CLIP demonstrates greater robustness to data scarcity, particularly at extreme reductions ($\frac{1}{50}$). Shaded regions indicate 80\% confidence intervals.}
    \label{fig:few_shot_acc}
\end{figure}

As shown in Figure ~\ref{fig:few_shot_acc}, EEG-CLIP (green lines) maintains relatively stable performance across increasingly smaller  fractions of the training set, from $\frac{1}{2}$ down to $\frac{1}{50}$ of the original dataset. For pathology detection (top left), EEG-CLIP maintains strong performance even with minimal data ($\frac{1}{20}$), outperforming both baselines as data becomes extremely scarce. Age classification (top right) shows EEG-CLIP consistently outperforming other approaches across all data regimes. For gender and medication tasks (bottom panels), all models show performance degradation with reduced data, but EEG-CLIP demonstrates greater robustness to extreme data reductions ($\frac{1}{50}$).

Taken together, these quantitative results provide strong evidence for the quality and transferability of the multi-modal representations learned by EEG-CLIP. Performance across the range of evaluation paradigms demonstrates that EEG-CLIP successfully encodes general semantic relationships between EEG and text. This enables the model to generalize to new tasks and datasets without task-specific fine-tuning. The recent ELM-MIL approach by \cite{Gijsen2025} achieves superior performance (87.11\% balanced accuracy) through Multiple Instance Learning extensions that address fine-grained EEG-text alignment. While our approach achieves 84.7\%, both works demonstrate that multimodal language supervision significantly outperforms EEG-only baselines, validating this research direction. 

\subsection{Impact of the report sections on the representations}

To analyze how different report sections influence EEG representation learning, we conducted systematic ablation experiments by training separate models with single section inputs. Each model utilized identical methodology except for the textual input, which was restricted to specific report sections (impression, description, history, etc.). We also experimented with randomly sampling sub-strings from each section during training, but this approach led to decreased performance compared to using complete sections.

\begin{figure}[h]
\centering
\includegraphics[width=0.8\textwidth]{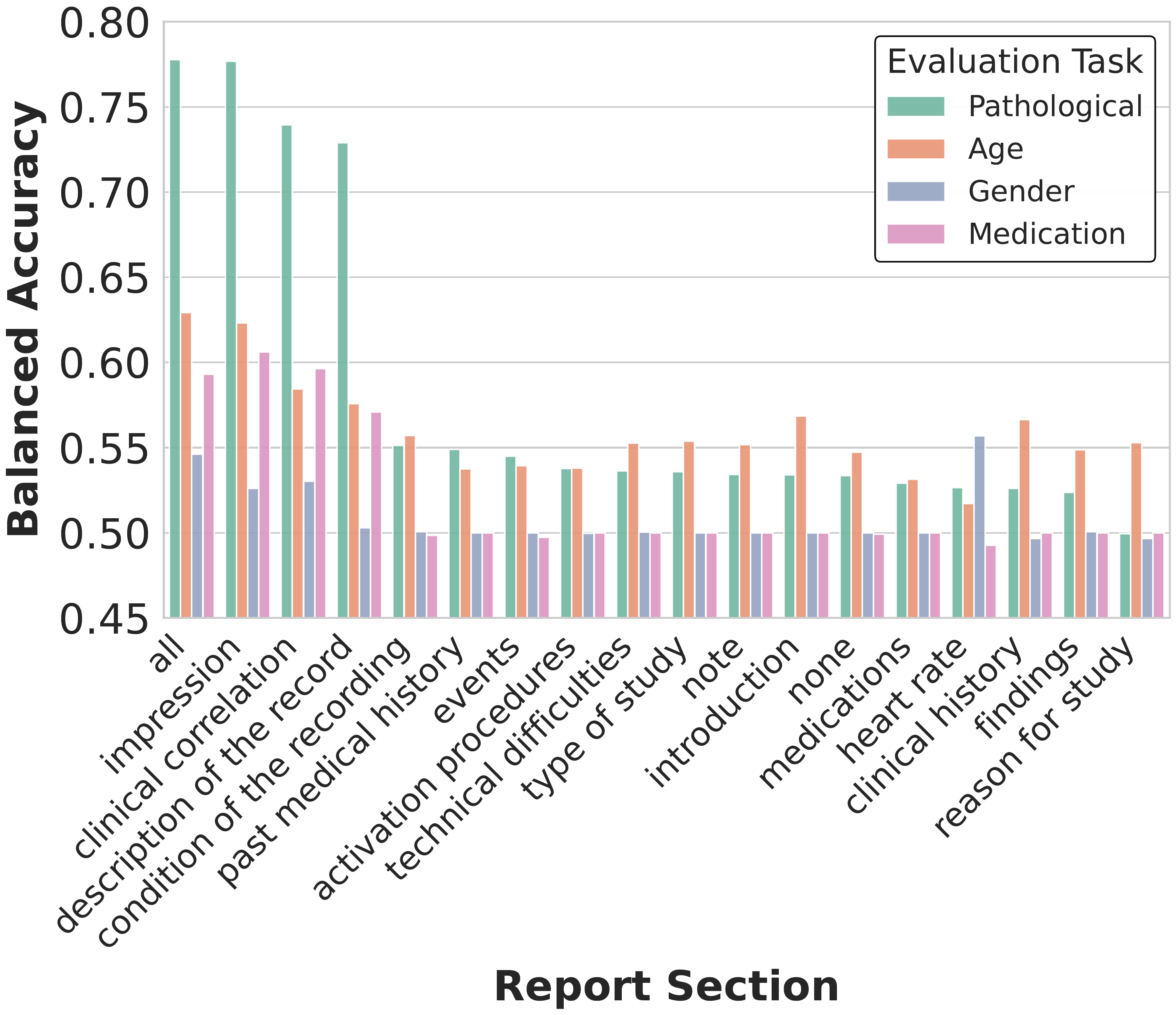} 
\caption{\textbf{Impact of report section selection on representation quality.} The heatmap visualization reveals how training on different clinical report sections affects decoding performance across tasks. While using all sections yields the best overall performance, specialized text categories show task-specific advantages. Notably, impressions and descriptions contribute most significantly to pathology detection, while sections like heart rate unexpectedly provide signal for gender classification, suggesting physiological correlations.}
\label{fig:hpo_category}
\end{figure}

As shown in Figure~\ref{fig:hpo_category}, while using all report sections yielded the best overall performance across tasks, certain section-specific models demonstrated unexpected strengths. Notably, the heart rate section model achieved superior accuracy in gender classification despite its brevity (average 2 words per report), suggesting cardiac pattern descriptions capture gender-specific physiological differences. Similarly, a model trained exclusively on technical difficulties notes showed enhanced sensitivity to pathological recordings, likely by learning to associate recording artifacts with abnormal brain activity.

The three primary sections (impression, description, and clinical history) provided the strongest individual contributions to performance, aligning with their higher word counts and prevalence across the dataset (Table~\ref{tuab-report-stats}). However, even sections with limited representation, such as medication lists, contributed unique predictive signals for specific tasks.

These findings reveal how specialized clinical descriptions, even when isolated, can help models detect task-relevant physiological patterns in EEG data. While combining all sections remains optimal for general-purpose representations, our analysis demonstrates the potential value of targeting specific report sections when developing specialized decoders or when working with incomplete clinical documentation.

\clearpage

\subsection{Study on parameter importance}

\subsubsection{Pre-training of the text encoder}

\begin{figure}[h]
\centering
\includegraphics[width=0.6\textwidth]{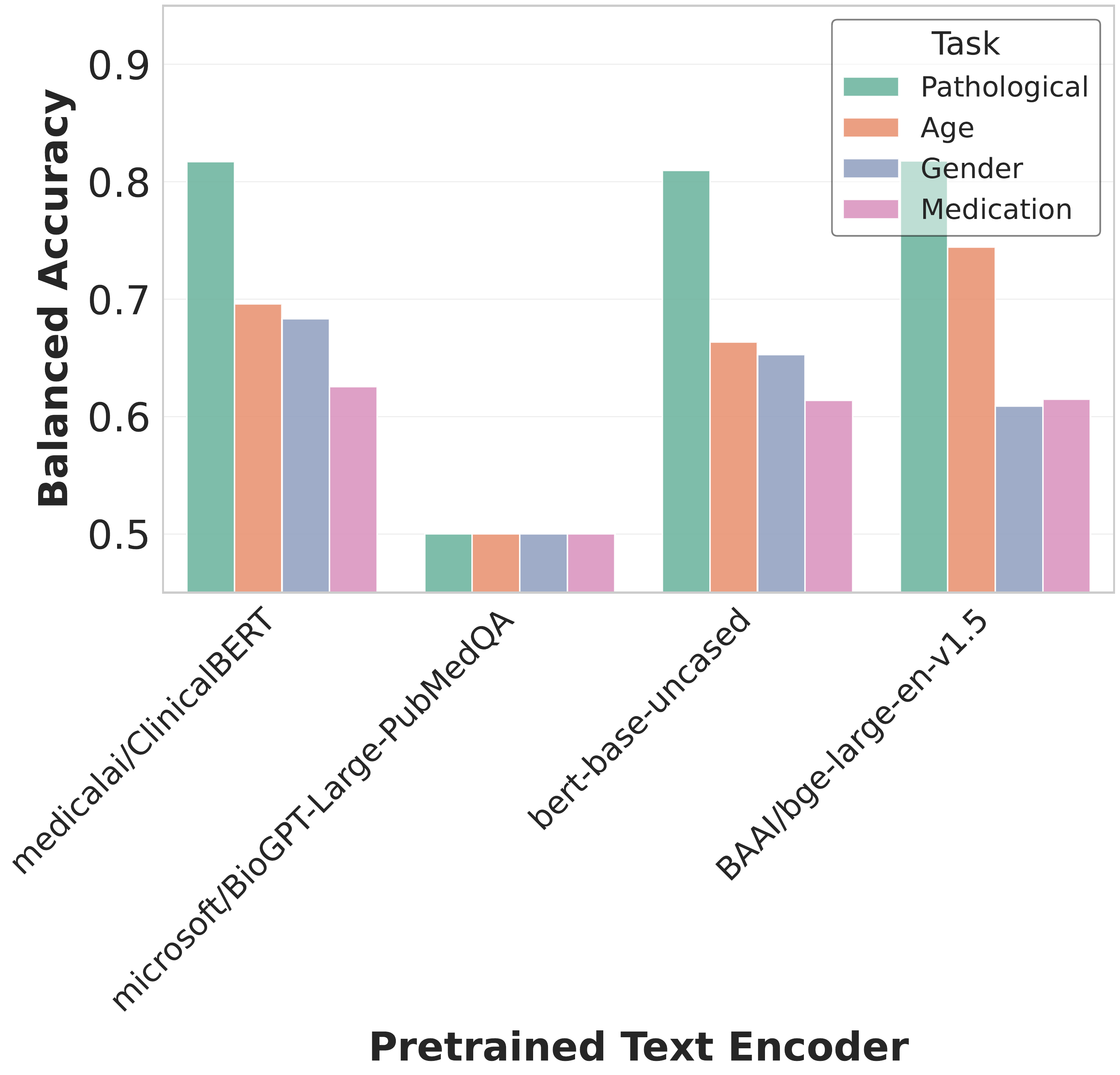} 
\caption{\textbf{Effect of text encoder selection on multimodal alignment.} The bar chart compares balanced accuracy across four decoding tasks using different text encoders. Medical/ClinicalBERT shows strong overall performance, while Microsoft/BioGPT-Large-PubMedQA performs poorly across all tasks. BERT-base-uncased maintains competitive pathology detection accuracy, and BAI/bge-large-en-v1.5 excels particularly at pathology and age classification. These results highlight how encoder choice significantly impacts cross-modal representation quality.}
\label{fig:hpo_encoder}
\end{figure}

We experimented with several variants of pre-trained language models as text encoders, available publicly on the Hugging Face Hub~\citep{wolf-etal-2020-transformers}, as shown in Figure~\ref{fig:hpo_encoder}. These included BERT-base-uncased~\citep{DBLP:journals/corr/abs-1810-04805}, a general domain model trained on Wikipedia and BookCorpus; ClinicalBERT~\citep{DBLP:journals/corr/abs-1904-05342}, a model finetuned on clinical notes; BioGPT-Large-PubMedQA~\citep{10.1093/bib/bbac409}, tailored for biomedical text; and BGE-Large~\citep{bge_embedding}, a model trained on scientific papers and designed for generation tasks.

ClinicalBERT demonstrated the strongest overall performance across tasks, highlighting the advantage of domain-specific pre-training for clinical text processing. Interestingly, the general-purpose BERT-base-uncased maintained competitive performance on pathology detection despite lacking medical specialization. BGE-Large showed particular strength in pathology and age classification tasks, while BioGPT-Large-PubMedQA consistently underperformed across all evaluations. These results emphasize how encoder architecture and pre-training domain significantly impact the quality of cross-modal representations in our EEG-text alignment framework.

\begin{figure}[h]
\centering
\includegraphics[width=0.6\textwidth]{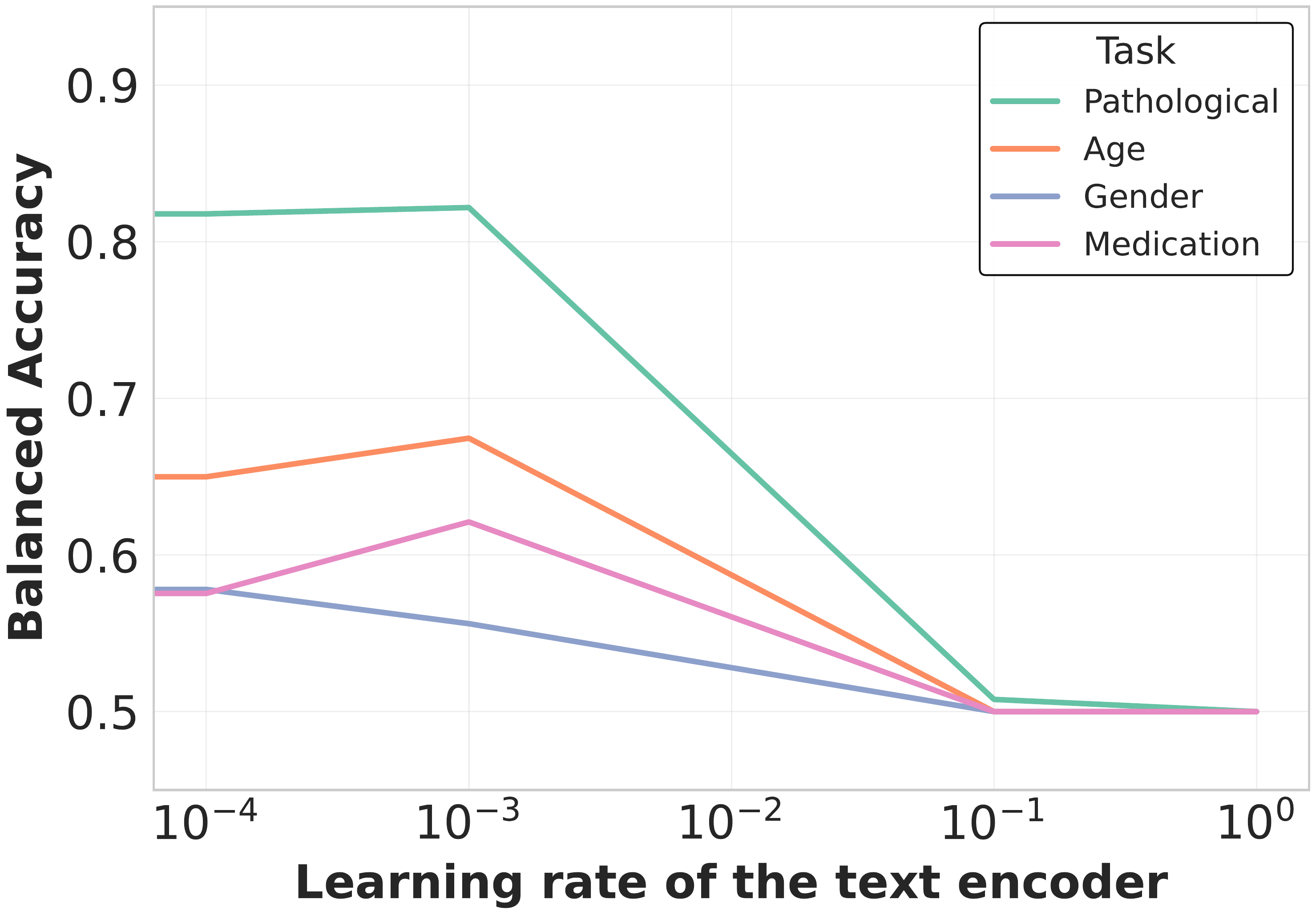} 
\caption{\textbf{Learning rate impact on text encoder fine-tuning.} The plot demonstrates how varying the learning rate ratio between the text encoder and the overall model affects representation quality. A moderate learning rate ($10^{-3}$ × the main learning rate) balances preservation of pretrained knowledge with adaptation to the EEG domain, optimizing cross-modal alignment without catastrophic forgetting of linguistic structures.}
\label{fig:hpo_lr}
\end{figure}

The learning rate ratio between the text encoder and EEG encoder also proved critical, as shown in Figure~\ref{fig:hpo_lr}. Optimal performance was achieved when the text encoder was fine-tuned at $10^{-3}$ times the learning rate of the EEG encoder, balancing adaptation of pre-trained linguistic knowledge while preserving domain-specific understanding.

\subsubsection{Projected embedding dimension}

\begin{figure}[h]
\centering
\includegraphics[width=0.6\textwidth]{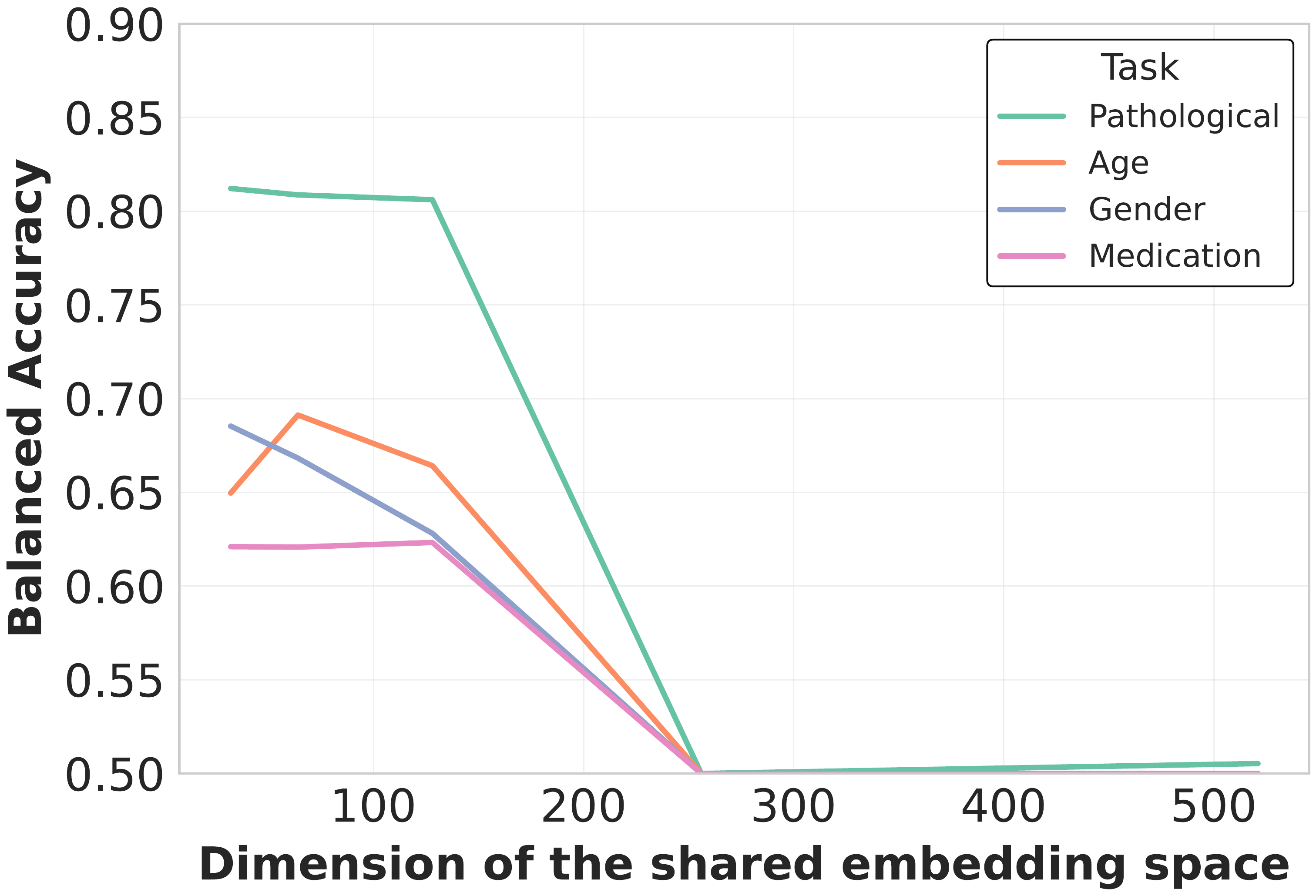} 
\caption{\textbf{Embedding dimensionality analysis for multimodal representation.} Counterintuitively, smaller embedding dimensions (32-128) consistently outperform larger ones (256-512) across all tasks. This suggests that compact shared embedding spaces better distill essential cross-modal patterns by enforcing more precise alignment between neurophysiological signals and their textual descriptions, while filtering out modality-specific noise.}
\label{fig:hpo_emb_size}
\end{figure}

\begin{figure}[h]
\centering
\includegraphics[width=0.6\textwidth]{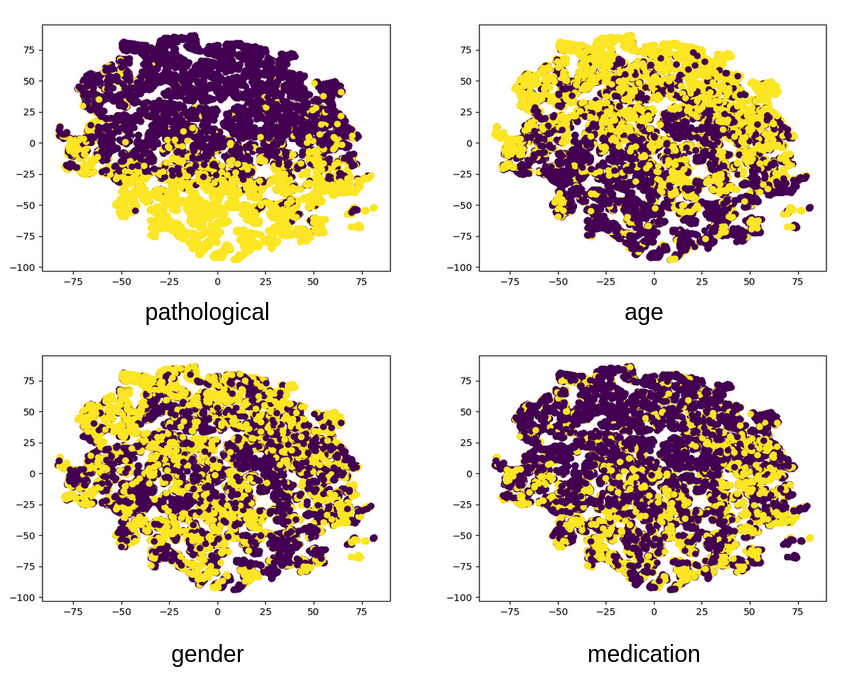} 
\caption{\textbf{Visualized embedding space using t-SNE dimensionality reduction.} The 2D projection of EEG embeddings from the evaluation set reveals clear clustering patterns corresponding to clinically relevant attributes. Pathological recordings (yellow) form distinct regions from normal recordings (purple), demonstrating that the unsupervised contrastive learning effectively captures diagnostically relevant features. Secondary clusters corresponding to age and gender are also visible, indicating the multifaceted nature of the learned representations.}
\label{fig:tsne_embeddings}
\end{figure}

Additionally, we analyzed EEG-CLIP model performance across varied hidden dimensionality sizes for the jointly learned EEG and text embeddings, as illustrated in Figure~\ref{fig:hpo_emb_size}. Counter to typical representation learning trends, we found higher decoding accuracy with smaller shared embedding spaces between 32-128 dimensions rather than larger 256 or 512 sizes. A t-SNE visualization of the 64-dimensional embeddings in Figure \ref{fig:tsne_embeddings} reveals clear clustering by pathological status, demonstrating effective semantic organization of the learned representation space.

This counterintuitive finding suggests that compressing both modalities into compact unified vectors distills patterns into their most essential characteristics necessary for generalization, while avoiding overfitting to training distribution artifacts that may occur in higher-dimensional spaces. The constrained dimensionality may also enforce more direct alignment between descriptive clinical concepts and underlying neurological patterns. These results indicate that EEG-CLIP benefits from lower-complexity manifolds that capture key cross-modal correspondences while filtering out extraneous signals.

\newpage

\subsection{Gradient Analysis for Model Interpretability}

In order to provide insights into EEG-CLIP's decision-making process, we performed gradient-based analysis to visualize which EEG regions contribute most strongly to embeddings aligned with specific textual concepts. We computed gradients of the cosine similarity between EEG embeddings and text embeddings for various prompts with regard to the EEG input that is forwarded through the EEG encoder. We computed those gradients with regard to the input in the frequency domain for easier analysis. Gradients were averaged across the validation set of TUAB.

\begin{figure}[h]
\centering
\includegraphics[width=0.8\textwidth]{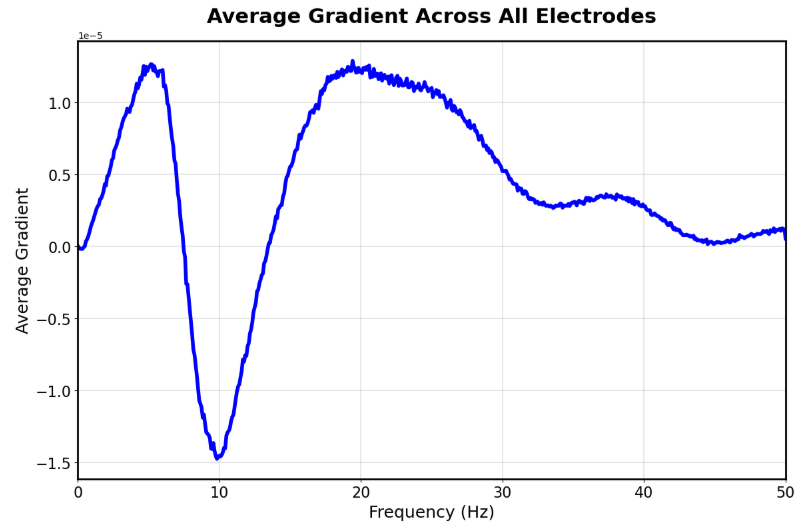}
\\[1em]
\includegraphics[width=0.8\textwidth]{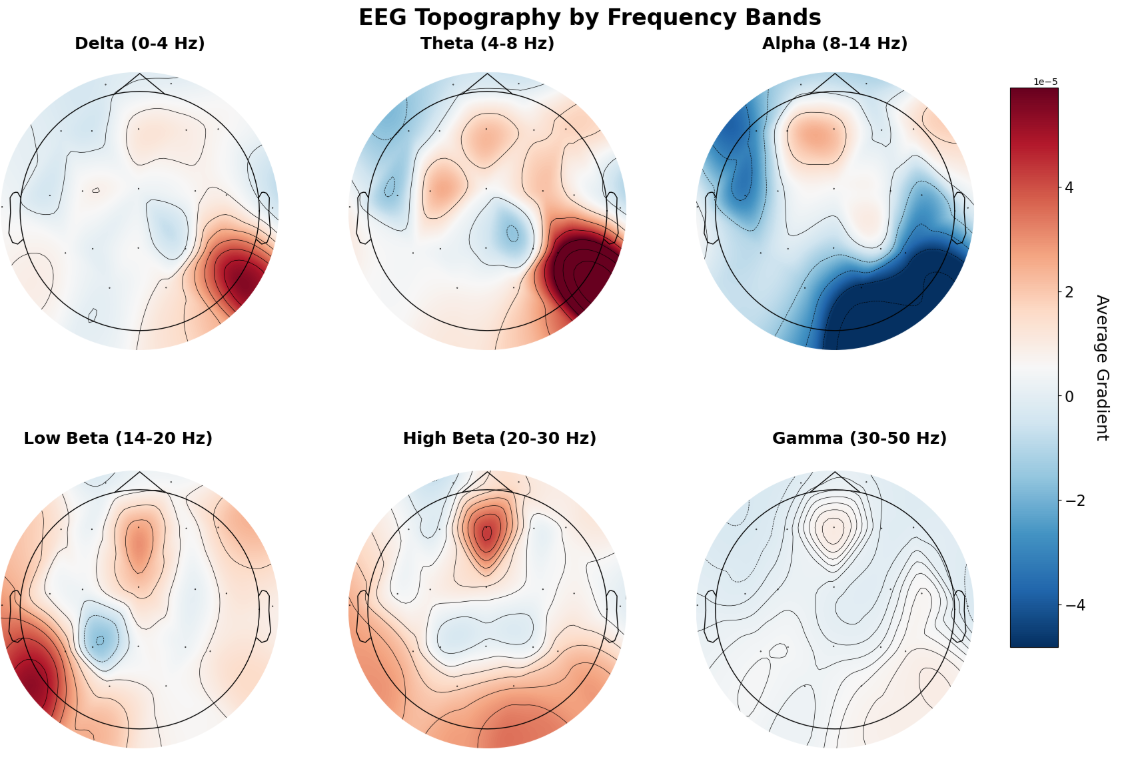}
\caption{\textbf{EEG-CLIP EEG encoder gradients that direct to the embeddings of "Excessive beta activity"} Top: EEG encoder gradients. Bottom: EEG encoder gradients per electrode. Gradients show elevated responses in the beta frequency range (14-30 Hz).}
\label{fig:gradients_excessive_beta}
\end{figure}

Analysis of prompts containing specific frequency descriptions reveals characteristic gradient patterns. The prompt "Excessive beta activity" (Figure~\ref{fig:gradients_excessive_beta}) produces elevated gradient responses in the beta frequency range (14-30 Hz).

\begin{figure}[h]
\centering
\includegraphics[width=0.8\textwidth]{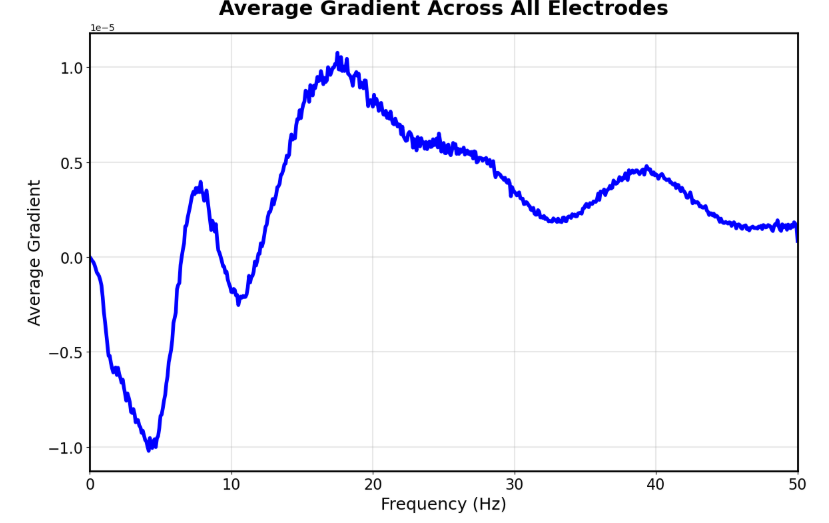}
\\[1em]
\includegraphics[width=0.8\textwidth]{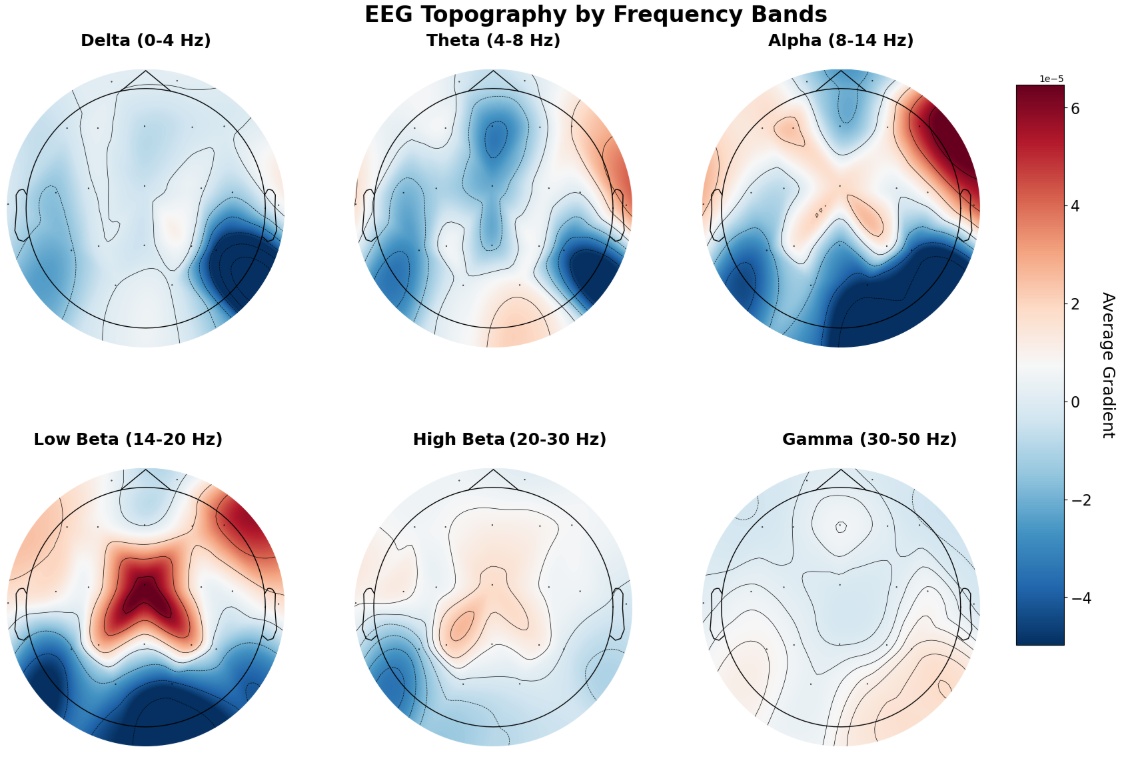}
\caption{\textbf{EEG-CLIP EEG encoder gradients that direct to the embeddings of "20 Hz beta spindles in central regions".} Top: EEG encoder gradients. Bottom: EEG encoder gradients per electrode. Gradients concentrate in the central electrode locations (C3, Cz, C4) with a peak at approximately 20 Hz, demonstrating the model's ability to capture both spatial and frequency-specific features mentioned in the text.}
\label{fig:gradients_central_beta}
\end{figure}

For the more spatially specific prompt "20 Hz beta spindles in central regions" (Figure~\ref{fig:gradients_central_beta}), gradients concentrate in the central electrode locations (C3, Cz, C4) with a peak at approximately 20 Hz, demonstrating the model's ability to capture both spatial and frequency-specific features mentioned in the text.

\begin{figure}[h]
\centering
\includegraphics[width=0.8\textwidth]{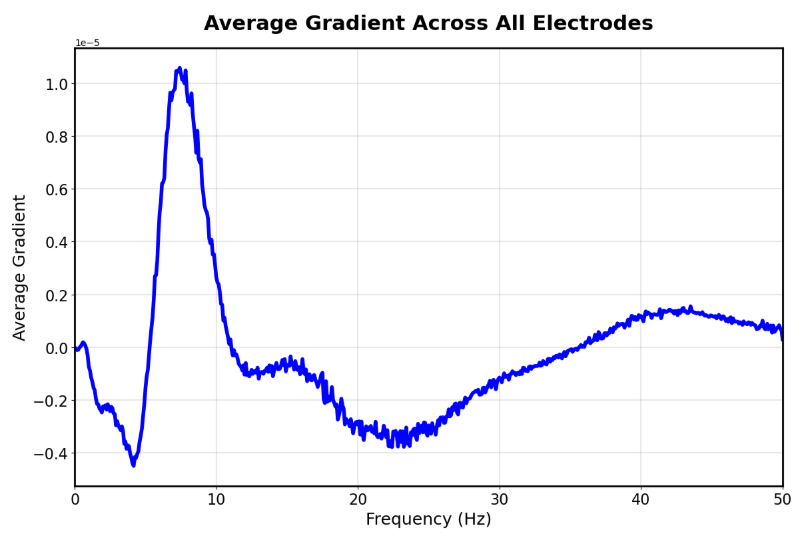}
\\[1em]
\includegraphics[width=0.8\textwidth]{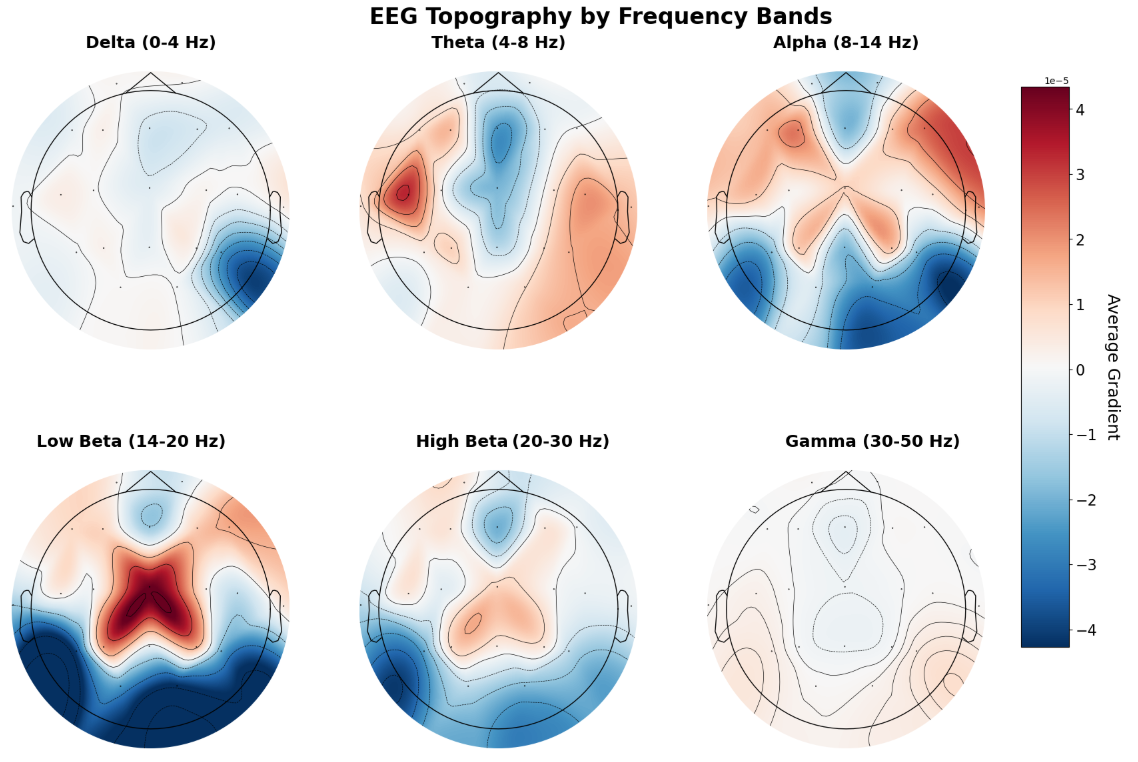}
\caption{\textbf{EEG-CLIP EEG encoder gradients that direct to the embeddings of "Left temporal sharp waves with a frequency of 6 Hz".} Top: EEG encoder gradients. Bottom: EEG encoder gradients per electrode. Higher magnitudes in left temporal electrodes (T3, T5) with a prominent peak around 6 Hz, indicating successful spatial-frequency alignment with the textual description.}
\label{fig:gradients_left_6hz}
\end{figure}

The prompt "Left temporal sharp waves with a frequency of 6 Hz" (Figure~\ref{fig:gradients_left_6hz}) produces gradients with lateralized patterns, showing higher magnitudes in left temporal electrodes (T3, T5) with a prominent peak around 6 Hz, indicating successful spatial-frequency alignment with the textual description.

\begin{figure}[h]
\centering
\includegraphics[width=0.8\textwidth]{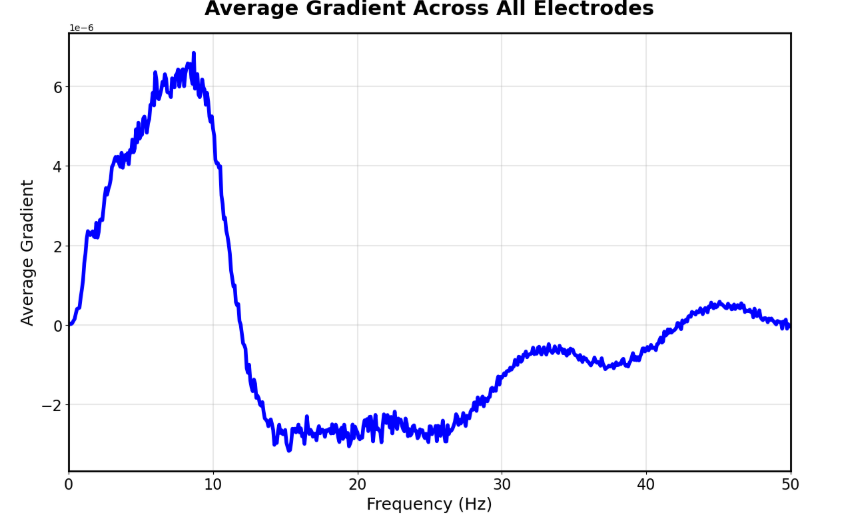}
\\[1em]
\includegraphics[width=0.8\textwidth]{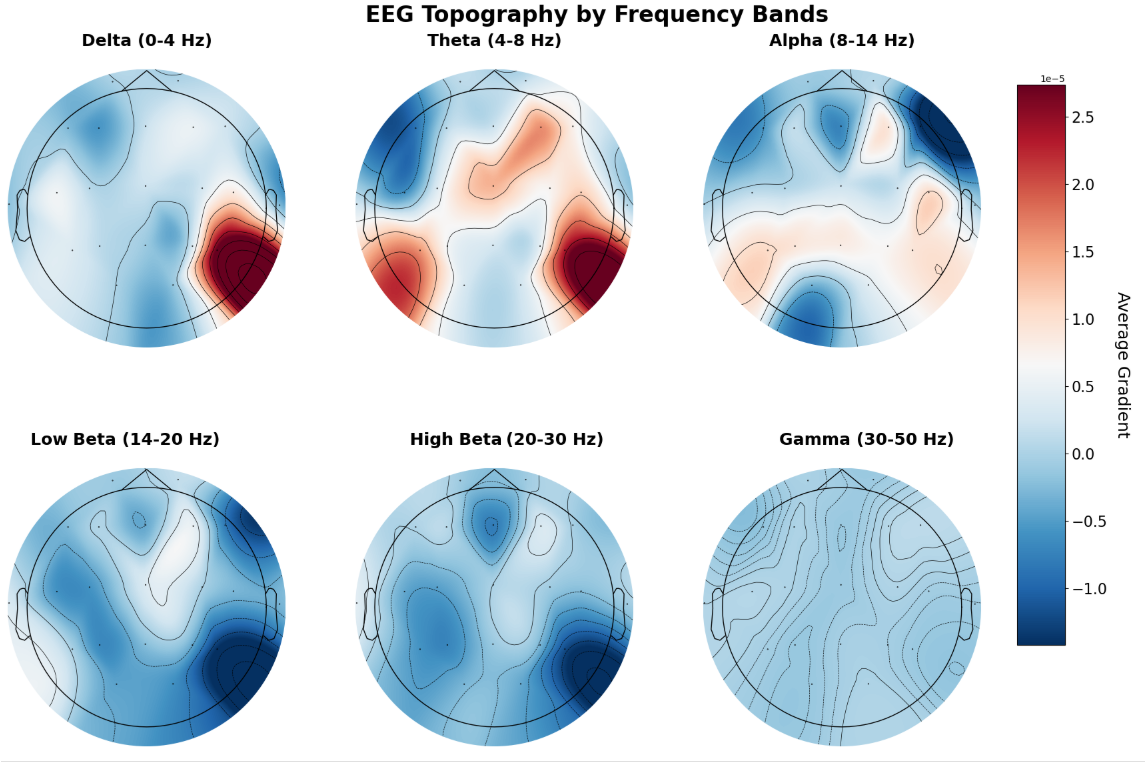}
\caption{\textbf{EEG-CLIP EEG encoder gradients that direct to the embeddings of "This is a normal recording".} Top: EEG encoder gradients. Bottom: EEG encoder gradients per electrode. Positive peak around 8-9 Hz in the alpha range, indicating the model identifies healthy recordings with increased alpha band activity.}
\label{fig:gradients_normal}
\end{figure}

Figure~\ref{fig:gradients_normal} shows the gradient patterns when the model aligns EEG signals with "This is a normal recording" embeddings. The frequency gradient analysis reveals a positive peak around 8-9 Hz in the alpha range, indicating the model identifies healthy recordings with increased alpha band activity. This pattern suggests the model has learned to identify normal EEG patterns through alpha frequency features.

These gradient visualizations provide preliminary evidence that EEG-CLIP learns clinically relevant spatial-temporal patterns rather than purely relying on spurious correlations. The distinct gradient patterns between different prompts suggest the model captures meaningful neurophysiological differences. However, more sophisticated interpretability methods and validation with clinical experts would be needed to fully understand the clinical relevance of these learned representations.

\newpage

\clearpage  

\newpage

\section{Discussion}

Our experiments demonstrate that EEG-CLIP successfully learns to align EEG recordings and their clinical text descriptions in a shared embedding space. This approach shows promise for developing more versatile and generalizable EEG representations that can transfer across multiple decoding tasks.

The model achieves strong performance on standard classification tasks (balanced accuracies of 0.847 for pathology, 0.702 for gender, and 0.747 for age) when using an MLP classifier head. Most notably, EEG-CLIP demonstrates zero-shot classification capabilities, achieving 0.755 balanced accuracy on pathology detection using only natural language prompts without any task-specific training. In low-data regimes, our approach shows substantial benefits over models trained from scratch or pretrained on alternative tasks, suggesting efficient capture of generalizable features.

Our ablation studies reveal that while using all report sections yields the best overall performance, specific sections provide distinct advantages for certain tasks. Interestingly, we found that smaller embedding dimensions (32-128) outperformed larger ones, contrary to common intuition in representation learning. This suggests that compressing information into a more compact shared embedding space may better distill essential cross-modal patterns.

Despite these promising results, several limitations warrant discussion. A significant limitation is our primary reliance on TUAB for evaluation, which raises valid concerns about dataset-specific biases and overfitting. Clinical reports in the dataset vary in quality, detail, and structure, potentially limiting the model's ability to learn precise alignments. Additionally, our current approach treats entire EEG recordings and their corresponding reports as aligned pairs, whereas more fine-grained temporal alignment between specific EEG segments and relevant report sections could improve performance, as demonstrated by recent Multiple Instance Learning frameworks \citep{Gijsen2025}. While we implemented validation strategies including held-out evaluation sets, cross-validation for few-shot experiments, and evaluation across four diverse tasks, broader multi-site validation across different hospital systems, recording protocols, patient populations, and clinical practices would strengthen generalizability claims.

Future work could explore methods for aligning specific EEG segments with relevant sentences in clinical reports, scaling to larger and more diverse EEG datasets, incorporating additional data modalities, and developing methods to interpret the learned representations in terms of clinically meaningful EEG patterns. The field would benefit from standardized multi-site evaluation protocols for EEG-language models to better assess generalizability across diverse clinical settings.

In conclusion, EEG-CLIP demonstrates the feasibility of contrastive learning between EEG signals and natural language descriptions for developing more general and transferable EEG representations. While this work represents an initial exploration, the approach opens up new possibilities for leveraging unstructured clinical text to enhance deep learning models for EEG analysis, potentially leading to more flexible and data-efficient tools for neurological assessment and research.

\newpage

\bibliographystyle{tmlr} 
\bibliography{main}

\begin{thebibliography}{14}
\providecommand{\natexlab}[1]{#1}
\providecommand{\url}[1]{\texttt{#1}}
\expandafter\ifx\csname urlstyle\endcsname\relax
  \providecommand{\doi}[1]{doi: #1}\else
  \providecommand{\doi}{doi: \begingroup \urlstyle{rm}\Url}\fi

\bibitem[Devlin et~al.(2018)Devlin, Chang, Lee, and Toutanova]{DBLP:journals/corr/abs-1810-04805}
Jacob Devlin, Ming{-}Wei Chang, Kenton Lee, and Kristina Toutanova.
\newblock {BERT:} pre-training of deep bidirectional transformers for language understanding.
\newblock \emph{CoRR}, abs/1810.04805, 2018.
\newblock URL \url{http://arxiv.org/abs/1810.04805}.

\bibitem[Devlin et~al.(2019)Devlin, Chang, Lee, and Toutanova]{devlin2019}
Jacob Devlin, Ming-Wei Chang, Kenton Lee, and Kristina Toutanova.
\newblock Bert: Pre-training of deep bidirectional transformers for language understanding.
\newblock In \emph{Proceedings of the 2019 conference of the North American chapter of the association for computational linguistics: human language technologies, volume 1 (long and short papers)}, pp.\  4171--4186, 2019.

\bibitem[Gijsen \& Ritter(2024)Gijsen and Ritter]{Gijsen2025}
Sam Gijsen and Kerstin Ritter.
\newblock Eeg-language modeling for pathology detection.
\newblock \emph{arXiv preprint arXiv:2409.07480}, 2024.

\bibitem[Heilmeyer et~al.(2018)Heilmeyer, Schirrmeister, Fiederer, Völker, Behncke, and Ball]{heilmeyer2018}
Felix~A. Heilmeyer, Robin~T. Schirrmeister, Lukas D.~J. Fiederer, Martin Völker, Joos Behncke, and Tonio Ball.
\newblock A large-scale evaluation framework for eeg deep learning architectures.
\newblock In \emph{2018 IEEE International Conference on Systems, Man, and Cybernetics (SMC)}, pp.\  1039--1045, 2018.
\newblock \doi{10.1109/SMC.2018.00185}.
\newblock URL \url{http://arxiv.org/abs/1806.07741}.

\bibitem[Huang et~al.(2019)Huang, Altosaar, and Ranganath]{DBLP:journals/corr/abs-1904-05342}
Kexin Huang, Jaan Altosaar, and Rajesh Ranganath.
\newblock Clinicalbert: Modeling clinical notes and predicting hospital readmission.
\newblock \emph{CoRR}, abs/1904.05342, 2019.
\newblock URL \url{http://arxiv.org/abs/1904.05342}.

\bibitem[Khan et~al.(2018)Khan, Ghafoor, and Hong]{khan_early_2018}
M.~Jawad Khan, Usman Ghafoor, and Keum-Shik Hong.
\newblock Early {Detection} of {Hemodynamic} {Responses} {Using} {EEG}: {A} {Hybrid} {EEG}-{fNIRS} {Study}.
\newblock \emph{Frontiers in Human Neuroscience}, 12, November 2018.
\newblock ISSN 1662-5161.
\newblock \doi{10.3389/fnhum.2018.00479}.
\newblock URL \url{https://www.frontiersin.org/journals/human-neuroscience/articles/10.3389/fnhum.2018.00479/full}.
\newblock Publisher: Frontiers.

\bibitem[Luo et~al.(2022)Luo, Sun, Xia, Qin, Zhang, Poon, and Liu]{10.1093/bib/bbac409}
Renqian Luo, Liai Sun, Yingce Xia, Tao Qin, Sheng Zhang, Hoifung Poon, and Tie-Yan Liu.
\newblock {BioGPT: generative pre-trained transformer for biomedical text generation and mining}.
\newblock \emph{Briefings in Bioinformatics}, 23\penalty0 (6), 09 2022.
\newblock ISSN 1477-4054.
\newblock \doi{10.1093/bib/bbac409}.
\newblock URL \url{https://doi.org/10.1093/bib/bbac409}.
\newblock bbac409.

\bibitem[Lévy et~al.(2025)Lévy, Zhang, Pinet, Rapin, Banville, d'Ascoli, and King]{levy2025}
Jarod Lévy, Mingfang Zhang, Svetlana Pinet, Jérémy Rapin, Hubert Banville, Stéphane d'Ascoli, and Jean-Rémi King.
\newblock Brain-to-text decoding: A non-invasive approach via typing.
\newblock \emph{arXiv preprint}, 2025.
\newblock URL \url{https://arxiv.org/abs/2502.17480}.

\bibitem[Obeid \& Picone(2016)Obeid and Picone]{obeid2016}
Iyad Obeid and Joseph Picone.
\newblock The temple university hospital eeg data corpus.
\newblock \emph{Frontiers in Neuroscience}, 10:\penalty0 196, 2016.
\newblock \doi{10.3389/fnins.2016.00196}.
\newblock URL \url{https://www.ncbi.nlm.nih.gov/pmc/articles/PMC4865520/}.

\bibitem[Radford et~al.(2021)Radford, Kim, Hallacy, Ramesh, Goh, Agarwal, Sastry, Askell, Mishkin, Clark, et~al.]{radford2021}
Alec Radford, Jong~Wook Kim, Chris Hallacy, Aditya Ramesh, Gabriel Goh, Sandhini Agarwal, Girish Sastry, Amanda Askell, Pamela Mishkin, Jack Clark, et~al.
\newblock Learning transferable visual models from natural language supervision.
\newblock In \emph{International conference on machine learning}, pp.\  8748--8763. PmLR, 2021.

\bibitem[Roy et~al.(2019)Roy, Banville, Albuquerque, Gramfort, Falk, and Faubert]{roy2019}
Yannick Roy, Hubert Banville, Isabela Albuquerque, Alexandre Gramfort, Tiago~H. Falk, and Jocelyn Faubert.
\newblock Deep learning-based electroencephalography analysis: a systematic review.
\newblock \emph{Journal of Neural Engineering}, 16\penalty0 (5):\penalty0 051001, 2019.
\newblock \doi{10.1088/1741-2552/ab260c}.
\newblock URL \url{https://dx.doi.org/10.1088/1741-2552/ab260c}.

\bibitem[Schirrmeister et~al.(2017)Schirrmeister, Gemein, Eggensperger, Hutter, and Ball]{schirrmeister2018}
R.~Schirrmeister, L.~Gemein, K.~Eggensperger, F.~Hutter, and T.~Ball.
\newblock Deep learning with convolutional neural networks for decoding and visualization of {EEG} pathology.
\newblock In \emph{2017 {IEEE} {Signal} {Processing} in {Medicine} and {Biology} {Symposium} ({SPMB})}, pp.\  1--7, 2017.
\newblock \doi{10.1109/SPMB.2017.8257015}.

\bibitem[Wolf et~al.(2020)Wolf, Debut, Sanh, Chaumond, Delangue, Moi, Cistac, Rault, Louf, Funtowicz, Davison, Shleifer, von Platen, Ma, Jernite, Plu, Xu, Le~Scao, Gugger, Drame, Lhoest, and Rush]{wolf-etal-2020-transformers}
Thomas Wolf, Lysandre Debut, Victor Sanh, Julien Chaumond, Clement Delangue, Anthony Moi, Pierric Cistac, Tim Rault, Remi Louf, Morgan Funtowicz, Joe Davison, Sam Shleifer, Patrick von Platen, Clara Ma, Yacine Jernite, Julien Plu, Canwen Xu, Teven Le~Scao, Sylvain Gugger, Mariama Drame, Quentin Lhoest, and Alexander Rush.
\newblock Transformers: State-of-the-art natural language processing.
\newblock In Qun Liu and David Schlangen (eds.), \emph{Proceedings of the 2020 Conference on Empirical Methods in Natural Language Processing: System Demonstrations}, pp.\  38--45, Online, October 2020. Association for Computational Linguistics.
\newblock \doi{10.18653/v1/2020.emnlp-demos.6}.
\newblock URL \url{https://aclanthology.org/2020.emnlp-demos.6/}.

\bibitem[Xiao et~al.(2023)Xiao, Liu, Zhang, and Muennighoff]{bge_embedding}
Shitao Xiao, Zheng Liu, Peitian Zhang, and Niklas Muennighoff.
\newblock C-pack: Packaged resources to advance general chinese embedding, 2023.

\end{thebibliography}

\end{document}